\documentclass[10pt,twocolumn,letterpaper]{article}

\usepackage{iccv}
\usepackage{times}
\usepackage{epsfig}
\usepackage{graphicx}
\usepackage{amsmath}
\usepackage{amssymb}

\usepackage[font=small]{caption}
\usepackage{subcaption}
\usepackage[breaklinks=true,bookmarks=false]{hyperref}
\newcommand*\samethanks[1][\value{footnote}]{\footnotemark[#1]}
\iccvfinalcopy 

\def\iccvPaperID{****} 

\begin{document}

\title{Learning Person Trajectory Representations for Team Activity Analysis}

\author{Nazanin Mehrasa\thanks{equal contribution},  Yatao Zhong\samethanks,  Frederick Tung,  Luke Bornn,  Greg Mori\\
Simon Fraser University\\
{\tt\small\{nmehrasa, yataoz, ftung, lbornn\}@sfu.ca, mori@cs.sfu.ca}
}

\maketitle
\begin{abstract}
Activity analysis in which multiple people interact across a large space is challenging due to the interplay of individual actions and collective group dynamics.
We propose an end-to-end approach for learning person trajectory representations for group activity analysis. The learned representations encode rich spatio-temporal dependencies and capture useful motion patterns for recognizing individual events, as well as characteristic group dynamics that can be used to identify groups from their trajectories alone.
We develop our deep learning approach in the context of team sports, which provide well-defined sets of events (e.g. pass, shot) and groups of people (teams).
Analysis of events and team formations using NHL hockey and NBA basketball datasets demonstrate the generality of our approach.
\end{abstract}

\section{Introduction}

Human activity analysis is a fundamental problem in computer vision. 
Activities may involve a single person performing a series of actions to complete a task, or they can include multiple people distributed across a large space collectively trying to achieve a shared goal.
In this paper, we focus on the end-to-end learning of feature representations for analyzing activities involving multiple interacting people distributed in space. 
Our aim is to learn rich feature representations that encode useful information about individual events as well as the overall group dynamics.

\begin{figure}[t]
\centering
  \includegraphics[width=0.44\textwidth]{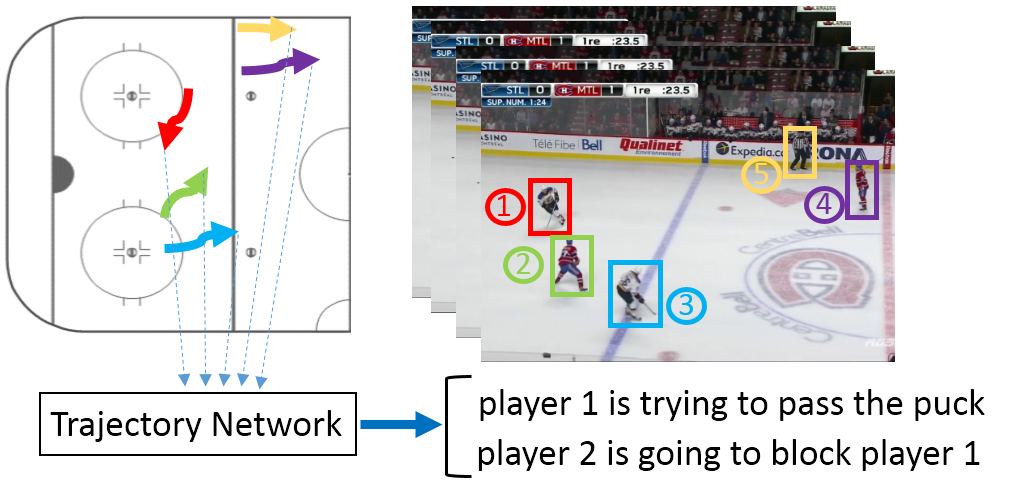}
\caption{When activities involve multiple people distributed in space, the relative trajectory patterns of different people can provide valuable cues for activity analysis. We learn rich trajectory representations that encode useful information for recognizing individual events as well as overall group dynamics in the context of team sports.}
  \label{fig:pull}
\end{figure}

We first observe that the trajectory a person takes while moving across a large space can provide valuable information as to the person's goal or intention. When activities involve multiple interacting people, the relative trajectory patterns of different people can also provide important interaction cues.
For example, consider the snapshot of time shown in Fig. \ref{fig:pull}, which shows a hockey player trying to pass the puck to a teammate while an opposing player moves to block the pass.   
A large volume of work has focused on visual features for recognizing individual actions.  These are typically built from challenging unconstrained Internet video datasets such as UCF Sports \cite{Rodriguez08actionmach}, UCF-101 \cite{soomro2012ucf101}, HMDB-51 \cite{Kuehne11},  and Sports-1M \cite{karpathy2014large}. 
However, it is hard to determine what a crowd of players is doing by simply inspecting the pixels inside the bounding boxes of players in a video clip. It is helpful to analyze the relative positions of players over time as well, so as to understand the spatio-temporal context and then understand the behavior of each player. 
We will show that learned person trajectory representations can encode useful individual and collective movement patterns for analyzing group actions in video.

We develop our deep learning approach in the context of team sports. Team sports are a useful testbed for complex group activity analysis as they provide a well defined set of events, such as pass, dump-in, and shot; well defined groups (teams) with distinct dynamics due to differences in strategy and tactics; and detailed annotations from sports analytics professionals.
At the same time, sports video analysis presents numerous challenges. Players move quickly and often frames are blurred due to this rapid movement. Thus, the input video clips do not always carry the rich visual information expected. Moreover, sports video, especially team sports, contains many player interactions. Interpreting those interactions can help understand their activities as a group, but the representations used to decode such interactions remains an open challenge. 

A body of literature focuses on group activity and human interaction \cite{amer2014hirf, choi2012unified, deng2016structure,DengZCLMRM15,ibrahim2015hierarchical,khamis2012combining, lan2012social, LanWYRM12,  Ramanathan_2013_CVPR}, some of which incorporate spatial information of individuals.  However, these representations tend to be hand-crafted and do not sufficiently encode the rich information of individual person movements and their interactions over time.

We propose a method to analyze team activities by using player locations in world coordinates over time.  
Analysis of trajectory data is not a new area of research. Many works have proposed computational methods to analyze player behaviors using trajectories. We refer readers to \cite{gudmundsson2016spatio} for a recent survey on this field of research. However, to the best of our knowledge, we are the first to learn features describing trajectories with convolutional neural networks, and the first to combine trajectory representations of multiple interacting actors in a single framework for group activity analysis. 

Our main contribution is a novel deep learning approach for group behavior analysis, which encodes spatio-temporal dependencies from person trajectories. To demonstrate the generality of our model, we analyze events and team formations using NHL (hockey) and NBA (basketball) datasets. We show that our learned representations capture useful information for recognizing events in team sports video. Our learned representations also capture distinctive group dynamics: we show that the collective trajectory patterns of a group can provide strong cues for identifying the group. 


\section{Related Work}

The existing literature on analyzing human activities is extensive.  Thorough surveys of earlier work include Gavrila~\cite{gavrila99visual} and Weinland et al.~\cite{weinland2011survey}.  Below, we review closely related work in activity recognition, including individual actions, group multi-person activities, and trajectory analysis.

\noindent {\bf Deep Learning for Action Recognition}:
Recently, deep learning has been brought to bear on the problem of action recognition.  Approaches for video-based action recognition include the two-stream network of Simonyan and Zisserman~\cite{simonyan2014two}, which fuse motion and appearance feature branches into a single network. Karpathy et al.~\cite{karpathy2014large} did extensive experiments on when and how to fuse information extracted from video frames. Donahue et al.~\cite{donahue2015long} extract features from each frame and encode temporal information using a recurrent neural net (LSTM \cite{hochreiter1997long}) for action recognition. Tran et al.~\cite{tran2015learning} extended traditional 2D convolution to the 3D case, where filters are applied to the spatial dimensions and temporal dimension simultaneously. These approaches have surpassed the performance of the best hand-crafted features, based on dense trajectories and motion features~\cite{wang:hal-00725627}.

\noindent {\bf Group Activity Recognition}:
Group activity recognition examines classifying the behaviour of multiple, interacting people.  Effective models typically  consider both individual actions and person-level interactions within the group. Previous works use hand-crafted features and model interactions with graphical models. Choi et al.~\cite{choi2009they} build hand-crafted descriptors of relative human poses. Lan et al.~\cite{LanWYRM12} and Amer et al.~\cite{amer2014hirf} utilize hierarchical models to understand collective activity among a group of people at different levels, ranging from atomic individual action to group activity in the scene.   The concept of social roles performed by people during interactions has also been studied~\cite{lan2012social, Ramanathan_2013_CVPR}.  All of these methods use hand-crafted representations of inter-person relationships.

Another line of work introduces structures into deep learning frameworks by integrating neural networks and graphical models in a unified framework \cite{ross2011learning, schwing2015fully, zheng2015conditional}. For example, Deng et al.~\cite{deng2016structure, DengZCLMRM15} apply deep structured models to collective activity recognition, learning dependencies between the actions of people in a scene.
However, these works do not consider spatio-temporal relationships between participants, which we believe would provide strong indication about how a group activity is formulated. Thus, we propose a model to incorporate spatial information by learning the dynamics of trajectories of each participant as well as their relative movements.  

The incorporation of track-level features as extra cues for interaction modeling was done by Choi et al.~\cite{choi2012unified} and Khamis et al.~\cite{khamis2012combining}.  
Recent work has developed more sophisticated deep temporal models for activity analysis.  Ramanathan et al.~\cite{ramanathan_cvpr16} utilize attention models to focus on key players in sports activties. Ibrahim et al.~\cite{ibrahim2015hierarchical} build hieararchical LSTMs to model multiple interacting people over time.  In contrast with this, our work learns trajectory features directly from human position inputs.

\noindent {\bf Trajectory Data Analytics}:
There exists significant literature on trajectory analysis focusing on team sports, such as basketball, soccer, and hockey. Applications within sports analytics include analyzing player and team performance, and mining underlying patterns that lead to certain results. Work in this field has included various statistical models to capture the spatio-temporal dynamics in player trajectories. We refer readers to a recent survey \cite{gudmundsson2016spatio} on detailed team sports analysis with trajectory data. 

Classic examples in the vision literature include Intille and Bobick~\cite{IntilleB01} who analyzed American football plays based on trajectory inputs.  M\'edioni et al.~\cite{MedioniCBHN01} utilized relative positioning between key elements in a scene, such as vehicles and checkpoints, to recognize activities.

In the sports context, Lucey et al.~\cite{LuceyBCMMS13} use a basis representation of person trajectories that utilizes roles.
The work that is most similar to ours is Wang and Zemel~\cite{wangclassifying}, which uses a traditional CNN plus RNN to classify NBA offensive patterns.  Person trajectories are converted to an image representation for input to the CNN. However, an image representation of trajectories is not an effective coding mechanism in that time is ignored, and further most pixels are zeros, making this representation redundant. We instead propose to build a neural network directly on top of raw trajectories.  

\begin{figure*}[t]
  \includegraphics[width=1.0\textwidth]{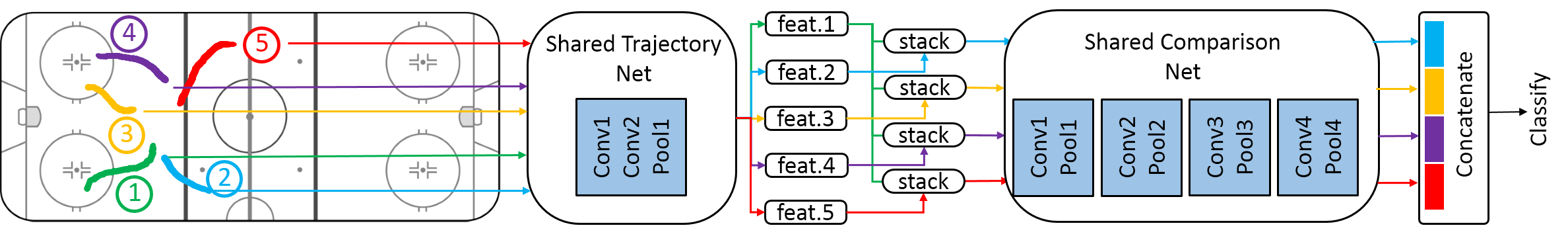}
  \caption{Our shared-compare trajectory network (Sec. \ref{sec:sharedcompare}) learns generic individual representations as well as features describing pairwise interactions. Pairs are determined by an ordering based on proximity to a key player. The shared-compare structure is useful for recognizing events such as passes and shots.}
  \label{fig:model}
\end{figure*}
\section{Proposed Approach}

We aim to learn trajectory representations that encode rich spatio-temporal patterns for recognizing individual events as well as characteristic group dynamics. Our proposed trajectory networks use layers of 1D temporal convolutions over person location inputs in world coordinates. We present two instantiations of this idea.

\begin{enumerate}
\item The \emph{shared} trajectory network takes a single person's trajectory as input. Since the weights of the network are shared across all people, the network learns a generic representation for any individual. The resulting individual representations are passed through a \emph{compare} network, which takes pairs of people as input, with pairs determined by an ordering based on proximity to a ``key" player. We show that the shared-compare network structure is useful for recognizing events such as passes and shots.
\item The \emph{stacked} trajectory network takes as input the stacked trajectories of all group members. This network learns representations that take into account all interactions among group members, and is effective at capturing overall group dynamics. We show that the stacked network structure can accurately predict group (team) identities by their member trajectories alone.
\end{enumerate}


\subsection{Shared-Compare Trajectory Network}
\label{sec:sharedcompare}

The shared-compare network structure is illustrated in Fig.\ \ref{fig:model}. The input to the shared network is a sequence of person world coordinates in the form of $\left(x_t, y_t\right)$, where $t$ is the frame number, associated with a single person. These inputs are obtained via state of the art tracking and camera calibration systems, which provide reasonably accurate, though sometimes noisy, data.  To learn the space-time variations in person trajectories, we propose to use layers of 1D convolutions. The network is ``shared" in that the same 1D convolutions are applied to each person.


Recall that a person trajectory is essentially a continuous signal.  We propose a direct way of interpreting a trajectory.  A 2D trajectory in world coordinates (e.g. player position in court or rink coordinates) has two separate continuous signals, one for the $x$ series and one for the $y$ series. We can split the input $\left[ (x_1, y_1), (x_2, y_2), \cdots, (x_T, y_T)\right]$ into two sequences $\left[x_1, x_2, \cdots, x_T\right]$ and $\left[y_1, y_2, \cdots, y_T\right]$, each being a 1D continuous signal. In our approach we treat these two sequences as two channels.  We build a convolutional neural network on top of these inputs, with 1D convolution operating on each input. By stacking layers of 1D convolution, we can learn combinations of $x$ and $y$ movements that are indicative of particular action classes.

In detail, let $X \in \mathbb{R}^{N\times T}$ denote the input, $F \in \mathbb{R}^{N \times W \times M}$ denote the filters in a convolutional layer and $O \in \mathbb{R}^{M\times T}$ denote the output, where $N$ is the number of input channels, $T$ is the length of input sequence, $W$ is the filter size and $M$ is the number of filters. To model the behaviour of a convolutional layer\footnote{We choose a step size of 1 when doing convolution and pad zeros to the input if the domain of a filter goes outside of the input.}, we perform the basic operation as follows:
\begin{equation} \label{eq:1}
    O_{k, t}=\sigma (\sum_{i=1}^{N} \sum_{j=1}^{W} X_{i, t+j-1} F_{i, j, k}).
\end{equation}

In the above formula, $\sigma \left(\cdot \right)$ can be any activation function. In our case, we choose ReLU for all activations. Two convolutional layers are followed by a max pooling layer to make the model shift-invariant and help reduce the dimension of the output. 

Let $Z \in \mathbb{R}^{M \times \lceil \frac{T}{S} \rceil}$ be the output of max pooling, where $S$ is the step size in the pooling operation, then we have
\begin{equation} \label{eq:2}
    Z_{k, t}=\max_{1\leq j\leq S}O_{k,\ (t-1)\cdot S+j}.
\end{equation}
To build a network with stacked convolutional and max pooling layers, we use the output $Z^{l-1}$ at layer $l-1$ as the input $X^{l}$ at layer $l$:
\begin{equation} \label{eq:3}
    X^l=Z^{l-1}.
\end{equation}
 We repeat the process described in Eq.~\ref{eq:1} and Eq.~\ref{eq:2} for a number of layers. To obtain the output of the shared trajectory network, we flatten the output of the last layer.

\begin{figure*}[t]
  \center
  \includegraphics[width=0.7\textwidth]{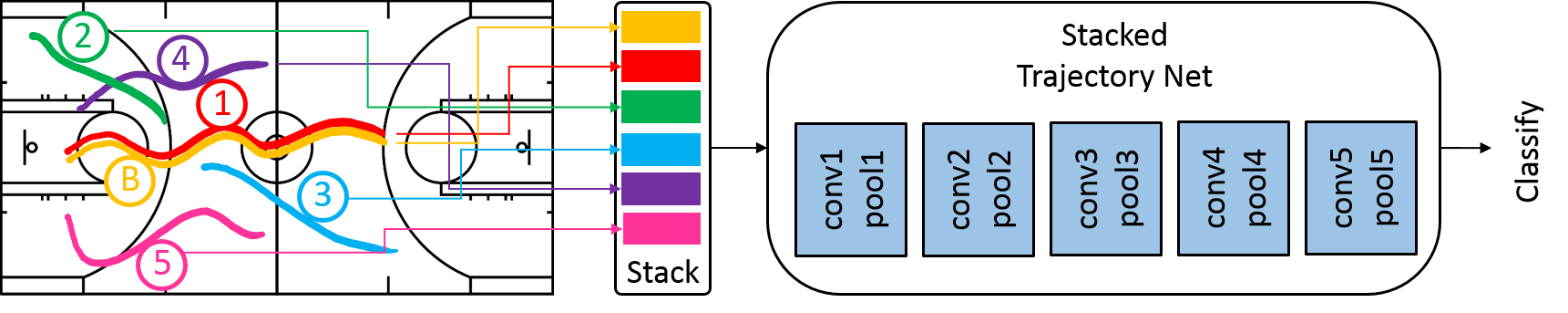}
  \caption{Our stacked trajectory network (Section \ref{sec:stacked}) takes trajectories of all group members as input and learns representations that consider all interactions within the group. The stacked network captures overall group dynamics and can be used to identify groups (teams) using member trajectories alone.}
  \label{fig:NBA}
\end{figure*}


The outputs of the shared network are then paired and passed to the compare network. The compare network aims to learn representations for the relative motion patterns of pairs of people. The comparison of trajectory pairs allows the network to learn interaction cues that can be useful for recognizing events or actions (e.g. pass, shot).

Pairs are formed relative to a ``key" person according to a pre-defined rule (we discuss the case of an unknown key person later). Denote the key person as $P^{(1)}$ and the other people as $P^{(2)}, P^{(3)}, ..., P^{({N_p})}$.
Given a pair $(P^{(1)},P^{(i)}), 2\leq i \leq N_p$, the individual trajectory representations of $P^{(1)}$ and $P^{(i)}$ (computed using the shared network) are input to a compare network consisting of several convolutional and max pooling layers. The output feature of the compare network can then be used as a learned feature representation of the trajectory pair. We apply the compare network to all pairs $(P^{(1)},P^{(i)}), 2\leq i \leq N_p$ relative to the key person, and concatenate the output features $Z^{(i)}$ to obtain the final feature descriptor of the group of people, $[Z^{(1)}, Z^{(2)}, \ldots , Z^{(N_p)}]$.
Next, a fully connected layer and softmax can be applied depending on the activity analysis task:

\begin{equation}
Z_e=\sigma([Z^{(1)}, Z^{(2)}, \ldots , Z^{(N_p)}] W_e)
\end{equation}

\noindent where $\sigma(\cdot)$ denotes softmax normalization, $F_e$ is the output feature size, $W_e \in \mathbb{R}^{(F \cdot N_p) \times F_e}$ are the weights of the fully connected layer, and $Z_e \in \mathbb{R}^{F_e}$ is the prediction vector which can be fed into a loss function for end-to-end training.

Note that when we perform the concatenation, we implicitly enforce an ordering among the group of people. Arbitrarily enforcing such an order is problematic. We set a consistent ordering by spatial proximity to the key person: $P^{(1)}$ is the key person, $P^{(2)}$ is the person closest to the key person, $P^{(3)}$ is the next closest person to the key person, and so on. 

The key person is often available in professional annotations for sports analytics, and typically corresponds to the player with possession of the ball or puck. However, the key person in a general non-sports setting may not be as well-defined. In addition, the key person may be unknown, or difficult to determine automatically, in general broadcast sports videos. When the key person is not provided, we adopt an average pooling strategy to determine $Z_e$. For each person in the scene, we fix that person as $P^{(1)}$, rank the other people accordingly based on proximity, and compute the network output. This produces $N^p$ prediction vectors $\{Z_e^{(1)}, Z_e^{(2)}, \ldots , Z_e^{(N_p)}\}$. We then apply element-wise average pooling over the $N^p$ prediction vectors to obtain the final prediction vector $Z^e$. Results for both known and unknown key player settings are presented in the experiments.

\subsection{Stacked Trajectory Network}
\label{sec:stacked}

The stacked trajectory network structure is illustrated in Fig.\ \ref{fig:NBA}. The input to this network is a stack of $(x_t,y_t)$ sequences for all members of the group (e.g. all members of the same team) and the ball. For example, suppose a group consists of $N_p$ people. Then the sequences $[(x^{(i)}_1, y^{(i)}_1), (x^{(i)}_2, y^{(i)}_2), \ldots , (x^{(i)}_T, y^{(i)}_T)]$ for each person $i$ and ball are stacked to form a $(C \cdot (N_p+1)) \times T$ dimensional input ($C=2$ for two channels $x$ and $y$. However, after stacking, its actual number of channels is $C \cdot (N_p+1)$). Similar to Section \ref{sec:sharedcompare}, a convolutional neural network consisting of layers of 1D convolutions is trained on this input. We set a consistent ordering by spatial proximity to the ball. The output of the final layer is flattened and input to a fully connected layer and softmax for task-specific end-to-end training.

By taking into account all combinations of interactions within the group, the stacked trajectory network learns overall group dynamics instead of individual representations.


\section{Datasets}

We conduct experiments on two datasets. The first includes player trajectories and visuals from broadcast video footage of NHL hockey games. The second consists of player trajectories extracted from an external tracking system recording player positions in NBA basketball games.

\subsection{The SPORTLOGiQ NHL Dataset}
\begin{figure}[h!]
  \center
  \includegraphics[width=0.4\textwidth]{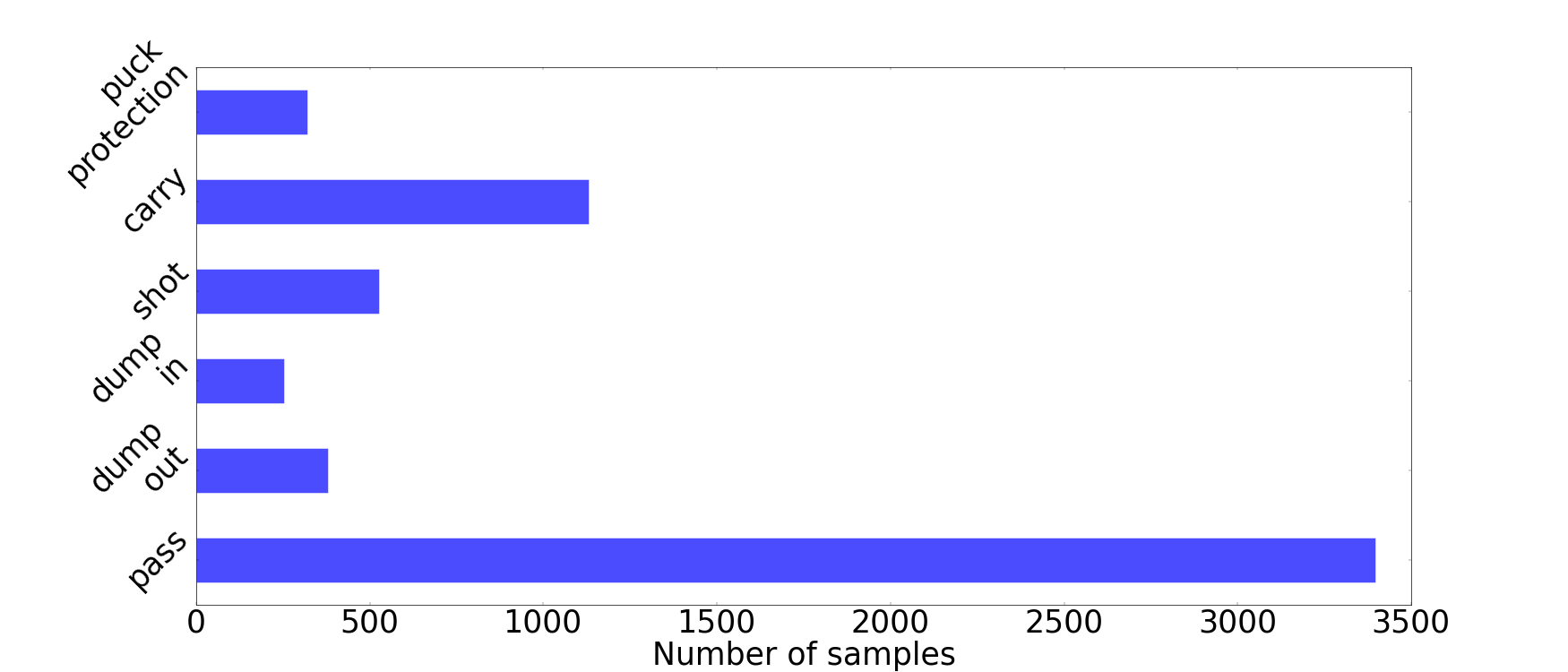}
  \caption{Number of samples per event in the SPORTLOGiQ NHL dataset.}
  \label{fig:nhl_hist}
\end{figure}

The SPORTLOGiQ NHL dataset includes both video and trajectory data. Unlike the NBA dataset where person trajectories are obtained from a multi-camera system, the player positions in the NHL dataset are estimated using a homography, which maps a pixel in image coordinates to a point in world coordinates. State of the art algorithms are used to automatically detect and track players in raw broadcast videos. If we have the bottom-middle point of a player bounding box, we can map this point to world coordinates with a homography matrix, hence acquiring the player position. The NHL dataset has detailed event annotation for each frame, each event being categorized into a super class and a fine-grained class. In our experiment, we use 8 games with 6 super classes: pass, dump out, dump in, shot, carry and puck protection. Fig.~\ref{fig:nhl_hist} shows the fraction of each event in the 8-game dataset. We will describe how we handle the event imbalance in Sec.~\ref{sec:evt_recog}.

\noindent {\bf Data Preprocessing}:
In a hockey game, typically there are 4 on-ice officials and 12 players (6 on each team). Thus, there can be at most 16 persons on the rink at the same time. In the following we do not make any distinction between officials and players and we use ``player" to refer to all people on the rink. Because the dataset is created from NHL broadcast videos where not all players are visible in each frame, we need to set a threshold $N_p$ so that our model can handle a fixed number of players. If the number of players available in a frame is less than $N_p$, we pad with zeros the part where players are unavailable. Each training sample consists of data from $N_p$ players. The data of each player includes a $T$-frame video clip (cropped from raw video using bounding boxes) and the corresponding $T$-frame trajectory estimated from this video clip. Note that our model supports variable-length input. If in some frames a player is not available, we set the data in these frames to zeros. In our experiments, $N_p$ is set to 5 and video frame size is set to $96 \times 96$. We set $T$ to 16 by first locating the center frame where an event happens and then cropping 7 frames before the center frame plus 8 frames after it. If the center frame of a certain event happens to be close to that of another event within 15 frames, we drop this sample. 

\subsection{The STATS SportVU NBA Dataset}

The STATS SportVU NBA dataset consists of real-time positions of players and the ball in 2D world coordinates captured by a six-camera system at a frame rate of 25 Hz. Each frame has complete annotations of the events happening in the frame, such as dribble, possession, shot, pass and rebound. The dataset we use has 1076 games during the 2013--2014 NBA season with approximately $10^6$ frames in each game.  We will use this dataset for evaluating the ability of our model to capture characteristic group dynamics. In particular, we attempt to predict the identity of a team using only the trajectories of its players during a game.

\begin{figure}[h!]
  \center
  \includegraphics[width=0.46\textwidth]{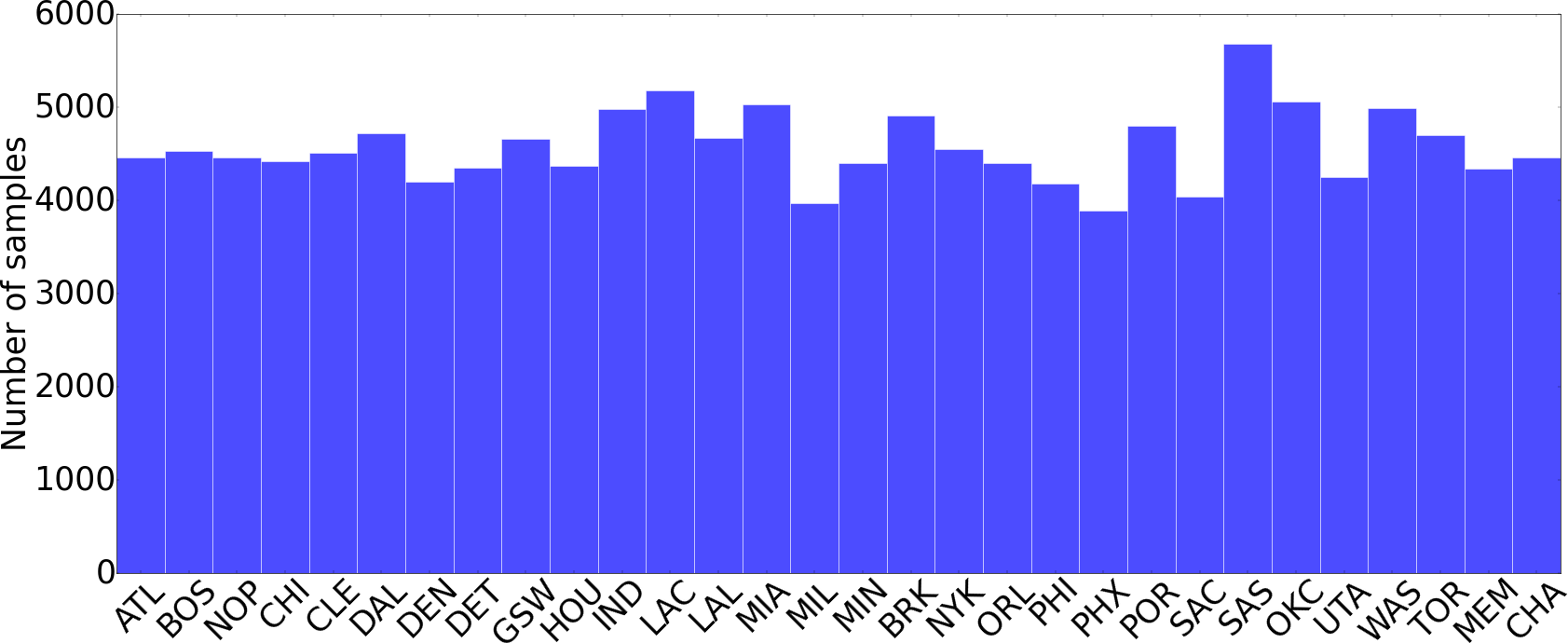}
  \caption{Number of possessions of each team in the STATS SportVU NBA dataset.} 
  \label{fig:nba_hist}
\end{figure}

\noindent {\bf Data Preprocessing}:
We extract 137176 possessions from the 1076 games for experiments. Each possession starts with an offensive team having possession of the ball and ends with a shot. We fix possession length to 200 frames. If a possession is longer than 200 frames, we crop it starting from the last frame and count the number of frames backward until it reaches 200. If a possession is shorter than 200 frames, we pad zeros to it. Originally there are 25 frames per second, but we sample only half of the frames in a second, so the sampled 200 frames actually represent a 16 second long sequence. There are in total 30 NBA teams. Fig.~\ref{fig:nba_hist} shows the number of possessions we extracted from each team in the dataset. We can see that this is a relatively balanced dataset, each team having a similar number of samples for experiments. 

\section{Experiments}
We conduct our experiments on the NHL and NBA datasets described above. To demonstrate that our shared-compare network is capable of learning temporal dynamics of person trajectories and group interactions, we perform event recognition using the NHL dataset. We then show that the stacked trajectory network captures overall group dynamics by performing team identity prediction on the NBA dataset, using only player trajectories.
\subsection{Event Recognition on the NHL Dataset}
\label{sec:evt_recog}
\noindent {\bf Task Description}:
The goal is to predict the event label given the trajectories of players on the rink. The events used are pass, dump out, dump in, shot, carry and puck protection. The number of samples of each event in the dataset is shown in Fig.~\ref{fig:nhl_hist}. This dataset is highly unbalanced with the pass event taking up half of the dataset. To resolve this problem, we minimize a weighted cross-entropy loss function during training. The weighting for each class is in inverse proportion to its frequency in the dataset.


\noindent {\bf Experiment Settings}:
We use the exact model shown in Fig.~\ref{fig:model} for the shared-compare model. The 1D convolutions of the shared trajectory network have 64 and 128 filters with a filter size of 3. The first two layers of compare network have 128 filters with filter size 3, and the last two have 256, 512  filters with filter size 3,2 respectively.  All max-pooling layers are performed with a window size of 2 with stride 2. The weights in the loss function for pass, dump out, dump in, shot, carry and puck protection are 0.07, 0.6, 1, 0.4, 0.2 and 0.7 respectively.
We use average precision as the evaluation metric.


\noindent {\bf Baselines}: We compare the proposed model with (1) a hand-crafted trajectory feature, IDT~\cite{wang:hal-00725627}; (2) deep learning using video features, C3D~\cite{tran2015learning}.  Comparisons with these baselines demonstrate that our learned trajectory features are better than previous trajectory descriptors; and our approach, which {\it only uses features from person trajectories}, can outperform methods that learn on top of video input. 
\begin{itemize}
    \item {\bf IDT}~\cite{wang:hal-00725627}: We use the same input data as in our method: sequences of $x,y$ coordinates of 5 players. We represent each trajectory using the IDT {\it Trajectory shape descriptor}.  An SVM with RBF kernel and  'one vs. rest' mechanism is deployed for multi-class classification.

    \item {\bf C3D}~\cite{tran2015learning}: We use a C3D model as the visual baseline, either trained from scratch using only the NHL dataset, or fine-tuned from a model pre-trained on Sports-1M. We obtain a short clip for each player by concatenating player bounding boxes across time. C3D takes each player clip as input. The weights of the network are shared across all players. Then the extracted features of players are concatenated using the same ordering as in our approach and passed to a softmax layer for classification.


\end{itemize}
\noindent {\bf Train phase}:
 We train all three networks by providing the key player, defined as the player who performs the action. The remaining players are ranked by proximity to the key player as described in Sec.~\ref{sec:sharedcompare}.
 
\noindent {\bf Test phase}: 
 For test time evaluation, we consider both the case where the key player is given, and the more general case where the key player is unknown. When the key player is not given, we adopt an average pooling strategy as described in Sec.~\ref{sec:sharedcompare}. In our experiments, we use 4 games for training, 2 games for validation and 2 games for testing.

\begin{figure*}[t]
  \includegraphics[width=1.0\textwidth]{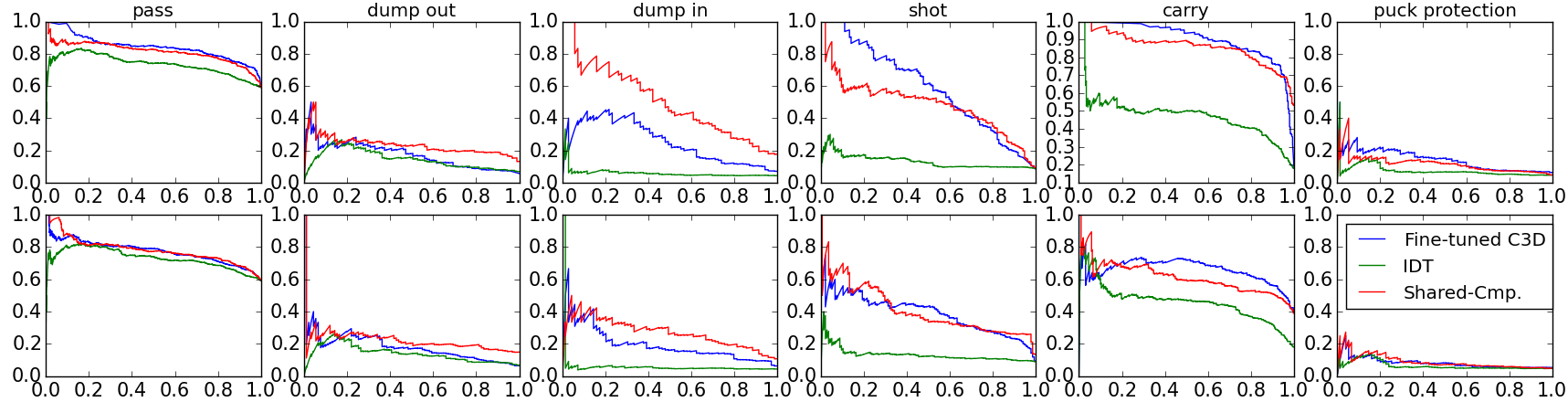}
  \caption{Precision-recall curves for each event in the NHL dataset. The first row shows the results of the known key player case while the second row the unknown key player case.}
  \label{fig:pr_curve}
\end{figure*}

\noindent {\bf Experiment Results}:
Tables \ref{table:unknown} and \ref{table:known} show experimental results when the key player is unknown and known, respectively. When the key player is unknown, our learned trajectory representations achieve 13.2\% higher mean average precision compared to state-of-the-art hand-crafted trajectory features (IDT). Compared to deep learning using video features, our shared-compare model obtains 8.7\% higher mAP when the C3D model is trained from scratch. Compared to the C3D baseline pre-trained on Sports-1M and fine-tuned on the NHL dataset, our shared-compare model obtains 1.7\% higher mAP while not requiring auxiliary training data (note that Sports-1M is a large sports dataset). Fig.~\ref{fig:pr_curve} shows precision-recall curves for each event. Our learned trajectory features consistently outperform the hand-crafted IDT trajectory features, and obtain mixed performance compared to the fine-tuned C3D model pre-trained on Sports-1M. 
When the key player is known, performance increases for all models; the shared-compare model improves by 10.5\% mAP. This result suggests that a useful direction for future work is the automatic prediction of the key person in the scene.

\begin{figure*}[t]
  \center
  \includegraphics[width=1.0\textwidth]{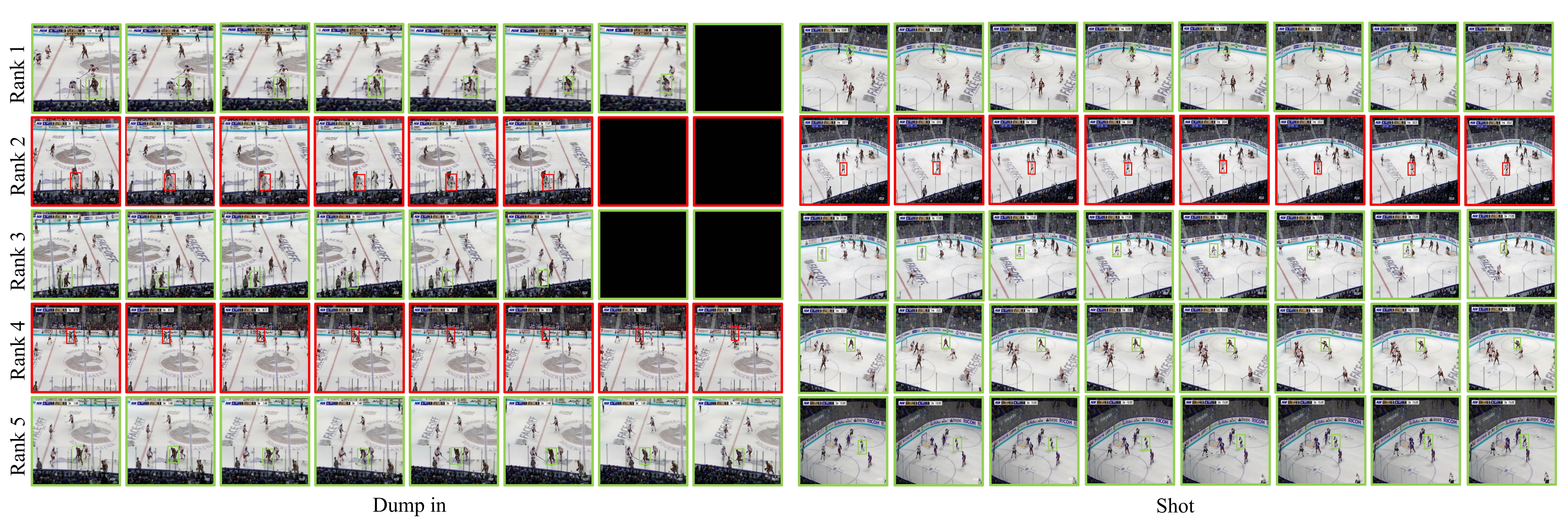}
  \caption{Top 5 candidates retrieved as dump in and shot respectively. Green sequences are true positives while red ones are false positives. The person with a bounding box is the ``key'' player who is performing the action. We show 8 frames of the 16-frame video clip by sub-sampling. If a frame is black, it means the key player is missing because of failure to detect and track the player. 
  }
  \label{fig:recall}
\end{figure*}

For the case where key player is provided, we visualize the top 5 candidates retrieved as dump in and shot in Fig. \ref{fig:recall}. The retrieved events look similar visually, highlighting the benefit of using the spatio-temporal relationships of the players to recognize events. 


\begin{table}
\small \addtolength{\tabcolsep}{-5pt}
\begin{center}
 \begin{tabular}{|c|r|r|r|r|}
 \hline
  & IDT & C3D & Fine-tuned C3D  & Shared-Cmp\\
 \hline
 pass & 72.86\%& 71.10\% &77.45\% & 78.13\% \\
 dump out & 13.75\%& 11.66\%&18.15\%  & 22.14\% \\
 dump in & 6.35\%& 7.58\% &19.04\% &  26.63\% \\
 shot & 13.05\%& 23.37\% &38.96\% & 40.52\% \\
 carry & 45.66\%& 64.75\% &65.65\% & 61.10\% \\
 puck protection & 6.28\%&6.50\% &7.98\% & 8.72\% \\
 \hline
 mAP & 26.32\%& 30.83\% &37.87\% &  39.54\% \\
 \hline
\end{tabular}
\caption{Average precision for event recognition on the NHL dataset, unknown key player.}
\label{table:unknown}
\end{center}
\end{table}

\begin{table}
\small \addtolength{\tabcolsep}{-5pt}
\begin{center}
 \begin{tabular}{|c|r|r|r|r|}
 \hline
  & IDT & C3D & 
  Fine-tuned C3D & Shared-Cmp\\
 \hline
 pass & 73.35\%& 77.30\% &84.34\% & 81.33\% \\
 dump out & 14.34\%& 10.17\% &17.10\% & 23.11\% \\
 dump in & 5.77\%& 10.25\% &24.83\% & 50.04\% \\
 shot &  13.07\%& 34.17\%  &58.88\% & 48.51\% \\
 carry &  47.38\%& 86.37\% &90.10\% & 85.96\% \\
 puck protection & 7.28\%& 11.83\%&13.99\%  & 11.54\% \\
 \hline
 mAP & 26.86\%& 38.35\% &48.21\% & 50.08\% \\
 \hline
\end{tabular}
\caption{Average precision for event recognition on the NHL dataset, known key player.}
\label{table:known}
\end{center}
\end{table}

\subsection{Team Identification on the NBA Dataset}

\noindent {\bf Task Description}: The goal is to predict the team identity given only its member trajectories. Good performance on this task demonstrates the capacity of our model to capture characteristic group dynamics.

\noindent {\bf Experiment Settings}:
We use the exact model shown in Fig.~\ref{fig:NBA} and described in Sec.~\ref{sec:stacked} for the stacked trajectory network. The $x$ and $y$ coordinates of the ball and 5 players are stacked together, resulting in a $200\times12$ matrix as input, where 200 is the length of the input sequence and 12 is the number of channels. We use 60\% of the 1076 games for training, 20\% for validation and 20\% for testing.

\noindent{\bf Baseline}:
We compare our model with IDT~\cite{wang:hal-00725627} . The same input data is used as in our method: sequences of $x,y$ coordinates of the ball and 5 players. Each trajectory is represented by the IDT {\it trajectory shape descriptor}. Then we concatenate the IDT features of the ball and the players and apply an SVM with RBF kernel and 'one vs. rest' mechanism for classification.

\noindent {\bf Measurement}:
We measure the performance of our model according to accuracy and hit-at-$k$ accuracy\footnote{Hit-at-$k$ accuracy: if any one of the top-$k$ predictions equals the ground truth label, we consider it as being correctly classified.} metrics, both of which are calculated over possessions. However, a single possession can hardly capture the distinctive group dynamics a team might possess. We therefore also aggregate all of a team's possessions in a game and predict the team identity using majority voting. 

\noindent {\bf Results and Analysis}:
Using the network architecture shown in Fig.~\ref{fig:NBA}, we are able to achieve 95\% accuracy in predicting a team's identity by majority voting over all its possessions in the game, given trajectory information alone.
We next explore the architecture of our model by varying the number of convolutional layers, the filter size and the number of filters in each layer. Tables~\ref{table:num_layers}, \ref{table:filter_sizes} and \ref{table:num_filters} show the results respectively. From Tables~\ref{table:num_layers} and \ref{table:num_filters}, we see that by increasing the number of layers and filters, generally we can obtain a more complex model to achieve better performance. However, as we increase the number of parameters in the model, we reach a limit that prevents us from acquiring further improvement by increasing the model complexity. For example, by adding two fully connected layers after the 5conv model in Table \ref{table:num_layers}, we obtain only a slight increase in possession-based accuracy and a drop in game-based accuracy. Also note that in Table~\ref{table:filter_sizes}, using small filter sizes generally leads to good results (see the first three models in Table~\ref{table:filter_sizes}). If we increase the filter size, we have a large decrease in model performance (see the last model in Table~\ref{table:filter_sizes}). Table~\ref{table:cmp_nba} compares our model with the IDT baseline and shows that our model learns distinctive group dynamics better than hand-crafted trajectory features. 
\begin{table}
\small
\begin{center}
 \begin{tabular}{|c|c|c|c|c|c|c|}
 \hline
  layers & acc & hit@2 & hit@3 & game acc \\
 \hline
  2conv & 10.68\% & 18.09\% & 24.31\% & 50.00\% \\ 
  3conv & 18.86\% &	28.89\%	& 36.47\% &	87.05\% \\
  4conv & 22.34\% & 33.03\% & 40.47\% & 93.41\% \\
  5conv & 24.78\% & 35.61\% & 42.95\% & 95.91\% \\
  5conv+2fc & 25.08\% &	35.83\% & 42.85\% & 94.32\% \\
 \hline
\end{tabular}
\caption{Metrics on models with different number of layers. All convolutional layers use a filter size of 3 except the first layer, where the filter size is 5. The number of filters in next layer is double the number in previous layer except the fifth layer (if any), where the number of filters is the same as that in the fourth layer. The number of neurons in fully connected layer is set to 1024.}
\label{table:num_layers}
\end{center}
\end{table}

\begin{table}
\small
\begin{center}
 \begin{tabular}{|c|c|c|c|c|c|c|}
 \hline
  filter sizes & acc & hit@2 & hit@3 & game acc \\
 \hline
  3 3 3 2 2 & 24.24\% & 35.36\% & 43.25\% & 94.10\% \\ 
  5 3 3 3 3 & 24.78\% & 35.61\% & 42.95\% & 95.91\% \\
  7 5 5 3 3 & 23.12\% & 33.48\% & 41.04\% & 95.45\% \\
  9 7 7 5 5 & 14.13\% & 23.15\% & 30.01\% & 62.05\% \\
 \hline
\end{tabular}
\caption{Metrics on models with different filter sizes. All models in the table use five convolutional layers with no fully connected layer. The filter sizes listed is in a bottom-up order and the number of filters used are 64, 128, 256, 512, 512 (bottom-up order). }
\label{table:filter_sizes}
\end{center}
\end{table}

\begin{table}
\small
\begin{center}
 \begin{tabular}{|c|c|c|c|c|c|c|}
 \hline
  base \# filters & acc & hit@2 & hit@3 & game acc \\
 \hline
  16 & 20.37\% & 30.71\% & 38.21\% & 81.14\% \\ 
  32 & 23.73\% & 34.55\% & 41.85\% & 92.95\% \\
  64 & 24.78\% & 35.61\% & 42.95\% & 95.91\% \\
  128 & 21.81\% & 32.10\% & 39.24\% & 94.45\% \\
 \hline
\end{tabular}
\caption{Metrics on models with different number of filters. All models in the table use five convolutional layers with no fully connected layer. The base number of filters listed in the table is the number of filters in the first layer. The number of filters in next layer is double the number in previous layer except that the fourth and the fifth layers have the same number of filters.}
\label{table:num_filters}
\end{center}
\end{table}

\begin{table}
\small
\begin{center}
 \begin{tabular}{|c|c|c|c|c|c|c|}
 \hline
  models & acc & game acc \\
 \hline
  IDT & 5.74\% & 9.10\% \\ 
  Stacked Traj. Net & 25.78\%  & 95.91\% \\
 \hline
\end{tabular}
\caption{Stacked trajectory network vs. IDT. Comparison with this baseline show that our model captures valuable collective features for understanding group dynamics}
\label{table:cmp_nba}
\end{center}
\end{table}

\begin{figure}[h]
 \center
\begin{subfigure}{0.43\columnwidth}
\includegraphics[width=0.84\columnwidth]{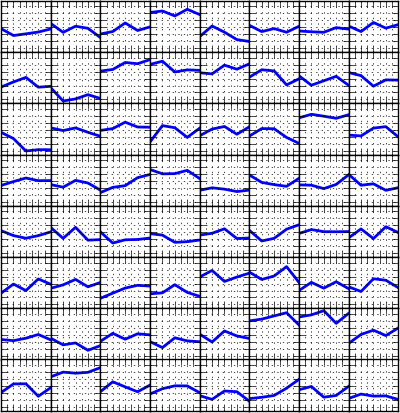} 
\caption{}
\label{fig:vis_filter}
\end{subfigure}
\begin{subfigure}{0.43\columnwidth}
\includegraphics[width=\columnwidth]{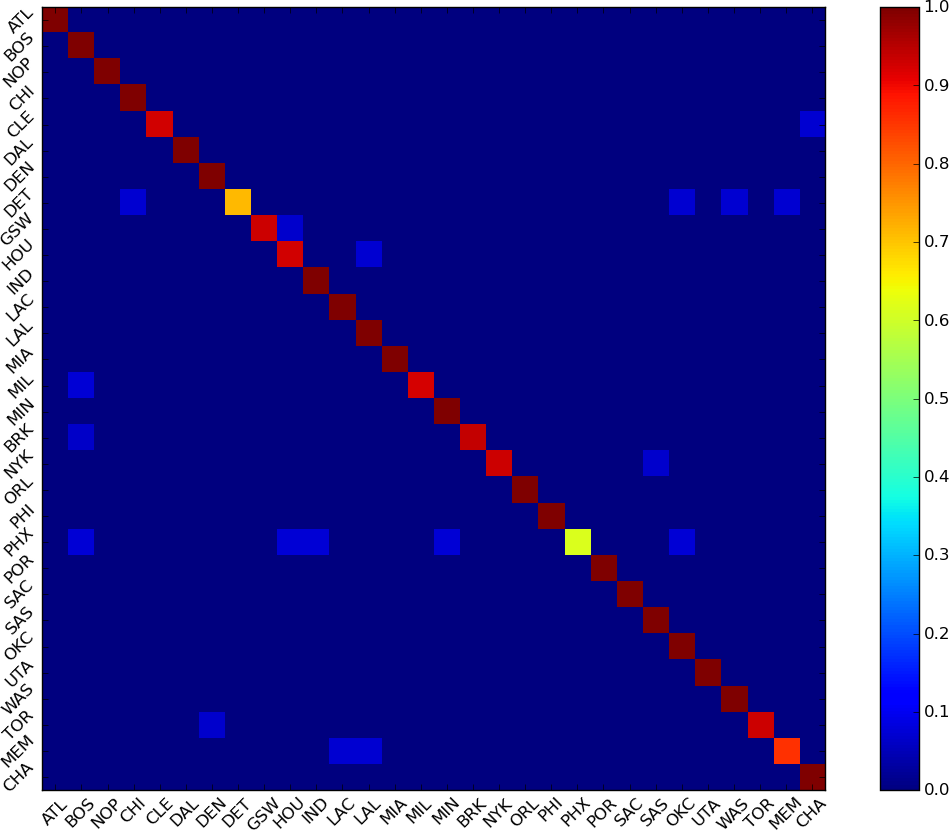}
\caption{}
\label{fig:confuse_matrix}
\end{subfigure}
 
\caption{(a) Visualization of the filters in the first convolutional layer. (b) Confusion matrix based on game-wise prediction. Both figures are created using the 5conv model in Table \ref{table:num_layers}.}
\label{fig:combined}
\end{figure}

Fig.~\ref{fig:confuse_matrix} shows the confusion matrix created from the 5conv model in Table \ref{table:num_layers}. For most teams, our model can correctly predict the team identity when aggregating over all of its possessions in a game. The worst case is the Phoenix Suns (PHX in Fig. \ref{fig:confuse_matrix}); the model has only a probability around 65\% to classify the Suns correctly, but this is still much better than chance performance.

To see what kind of patterns the model learns over the time dimension, we visualize a small fraction of the filters in the first convolutional layer. In Fig.~\ref{fig:vis_filter}, we show 64 filters learned from the input sequence of $x$ coordinates of the ball. Some of them appear to be ``Z" or ``S" shaped and some appear to be ``M" or ``W" shaped. Some of them are similar, so there could be redundancy in these filters. These temporal patterns are the building blocks that form discriminative representations to distinguish teams.

We also visualize in the supplementary video sample possessions that are confidently predicted as the Boston Celtics. Interestingly, we note that some of these possessions are a stereotypical Celtics offensive set play known as the pick-and-pop.


\section{Conclusion}


Team sports provide a useful testbed for understanding activities involving groups of people interacting to achieve shared goals.
We have presented deep neural network models for learning person trajectory representations for group activity analysis.
Our learned representations encode complex spatial-temporal dependencies for recognizing individual events, and capture distinctive group dynamics that can be used to identify teams given only member trajectories.

{\small
\bibliographystyle{ieee}
\bibliography{egbib}

\begin{thebibliography}{10}\itemsep=-1pt

\bibitem{amer2014hirf}
M.~R. Amer, P.~Lei, and S.~Todorovic.
\newblock Hirf: Hierarchical random field for collective activity recognition
  in videos.
\newblock In {\em European Conference on Computer Vision (ECCV)}, pages
  572--585. Springer, 2014.

\bibitem{choi2012unified}
W.~Choi and S.~Savarese.
\newblock A unified framework for multi-target tracking and collective activity
  recognition.
\newblock In {\em European Conference on Computer Vision (ECCV)}, pages
  215--230. Springer, 2012.

\bibitem{choi2009they}
W.~Choi, K.~Shahid, and S.~Savarese.
\newblock What are they doing?: Collective activity classification using
  spatio-temporal relationship among people.
\newblock In {\em Computer Vision Workshops (ICCV Workshops)}, pages
  1282--1289. IEEE, 2009.

\bibitem{deng2016structure}
Z.~Deng, A.~Vahdat, H.~Hu, and G.~Mori.
\newblock Structure inference machines: Recurrent neural networks for analyzing
  relations in group activity recognition.
\newblock In {\em Computer Vision and Pattern Recognition (CVPR)}, 2016.

\bibitem{DengZCLMRM15}
Z.~Deng, M.~Zhai, L.~Chen, Y.~Liu, S.~Muralidharan, M.~Roshtkhari, , and
  G.~Mori.
\newblock Deep structured models for group activity recognition.
\newblock In {\em British Machine Vision Conference (BMVC)}, 2015.

\bibitem{donahue2015long}
J.~Donahue, L.~Anne~Hendricks, S.~Guadarrama, M.~Rohrbach, S.~Venugopalan,
  K.~Saenko, and T.~Darrell.
\newblock Long-term recurrent convolutional networks for visual recognition and
  description.
\newblock In {\em Computer Vision and Pattern Recognition (CVPR)}, pages
  2625--2634, 2015.

\bibitem{gavrila99visual}
D.~M. Gavrila.
\newblock The visual analysis of human movement: {A} survey.
\newblock {\em Computer Vision and Image Understanding (CVIU)}, 73(1):82--98,
  1999.

\bibitem{gudmundsson2016spatio}
J.~Gudmundsson and M.~Horton.
\newblock Spatio-temporal analysis of team sports--a survey.
\newblock {\em arXiv preprint arXiv:1602.06994}, 2016.

\bibitem{hochreiter1997long}
S.~Hochreiter and J.~Schmidhuber.
\newblock Long short-term memory.
\newblock {\em Neural computation}, 9(8):1735--1780, 1997.

\bibitem{ibrahim2015hierarchical}
M.~Ibrahim, S.~Muralidharan, Z.~Deng, A.~Vahdat, and G.~Mori.
\newblock A hierarchical deep temporal model for group activity recognition.
\newblock In {\em Computer Vision and Pattern Recognition (CVPR)}, 2016.

\bibitem{IntilleB01}
S.~S. Intille and A.~Bobick.
\newblock Recognizing planned, multiperson action.
\newblock {\em Computer Vision and Image Understanding (CVIU)}, 81:414--445,
  2001.

\bibitem{karpathy2014large}
A.~Karpathy, G.~Toderici, S.~Shetty, T.~Leung, R.~Sukthankar, and L.~Fei-Fei.
\newblock Large-scale video classification with convolutional neural networks.
\newblock In {\em Computer Vision and Pattern Recognition (CVPR)}, pages
  1725--1732, 2014.

\bibitem{khamis2012combining}
S.~Khamis, V.~I. Morariu, and L.~S. Davis.
\newblock Combining per-frame and per-track cues for multi-person action
  recognition.
\newblock In {\em European Conference on Computer Vision (ECCV)}, pages
  116--129. Springer, 2012.

\bibitem{Kuehne11}
H.~Kuehne, H.~Jhuang, E.~Garrote, T.~Poggio, and T.~Serre.
\newblock {HMDB}: a large video database for human motion recognition.
\newblock In {\em The International Conference on Computer Vision (ICCV)},
  2011.

\bibitem{lan2012social}
T.~Lan, L.~Sigal, and G.~Mori.
\newblock Social roles in hierarchical models for human activity recognition.
\newblock In {\em Computer Vision and Pattern Recognition (CVPR)}, 2012.

\bibitem{LanWYRM12}
T.~Lan, Y.~Wang, W.~Yang, S.~Robinovitch, and G.~Mori.
\newblock Discriminative latent models for recognizing contextual group
  activities.
\newblock {\em IEEE Transactions on Pattern Analysis and Machine Intelligence
  (T-PAMI)}, 34(8):1549--1562, 2012.

\bibitem{LuceyBCMMS13}
P.~Lucey, A.~Bialkowski, G.~P.~K. Carr, S.~Morgan, I.~Matthews, and Y.~Sheikh.
\newblock Representing and discovering adversarial team behaviors using player
  roles.
\newblock In {\em Computer Vision and Pattern Recognition (CVPR)}, 2013.

\bibitem{MedioniCBHN01}
G.~M\'edioni, I.~Cohen, F.~Br\'emond, S.~Hongeng, and R.~Nevatia.
\newblock Event detection and analysis from video streams.
\newblock {\em IEEE Transactions on Pattern Analysis and Machine Intelligence
  (T-PAMI)}, 23(8):873--889, 2001.

\bibitem{ramanathan_cvpr16}
V.~Ramanathan, J.~Huang, S.~Abu-El-Haija, A.~Gorban, K.~Murphy, and L.~Fei-Fei.
\newblock Detecting events and key actors in multi-person videos.
\newblock In {\em Computer Vision and Pattern Recognition (CVPR)}, Las Vegas,
  USA, June 2016.

\bibitem{Ramanathan_2013_CVPR}
V.~Ramanathan, B.~Yao, and L.~Fei-Fei.
\newblock Social role discovery in human events.
\newblock In {\em Computer Vision and Pattern Recognition (CVPR)}, June 2013.

\bibitem{Rodriguez08actionmach}
M.~D. Rodriguez, J.~Ahmed, and M.~Shah.
\newblock Action mach: a spatio-temporal maximum average correlation height
  filter for action recognition.
\newblock In {\em Computer Vision and Pattern Recognition (CVPR)}, 2008.

\bibitem{ross2011learning}
S.~Ross, D.~Munoz, M.~Hebert, and J.~A. Bagnell.
\newblock Learning message-passing inference machines for structured
  prediction.
\newblock In {\em Computer Vision and Pattern Recognition (CVPR)}, pages
  2737--2744. IEEE, 2011.

\bibitem{schwing2015fully}
A.~G. Schwing and R.~Urtasun.
\newblock Fully connected deep structured networks.
\newblock {\em arXiv preprint arXiv:1503.02351}, 2015.

\bibitem{simonyan2014two}
K.~Simonyan and A.~Zisserman.
\newblock Two-stream convolutional networks for action recognition in videos.
\newblock In {\em Advances in Neural Information Processing Systems (NIPS)},
  pages 568--576, 2014.

\bibitem{soomro2012ucf101}
K.~Soomro, A.~R. Zamir, and M.~Shah.
\newblock Ucf101: A dataset of 101 human actions classes from videos in the
  wild.
\newblock {\em arXiv preprint arXiv:1212.0402}, 2012.

\bibitem{tran2015learning}
D.~Tran, L.~Bourdev, R.~Fergus, L.~Torresani, and M.~Paluri.
\newblock Learning spatiotemporal features with 3d convolutional networks.
\newblock In {\em International Conference on Computer Vision (ICCV)}, pages
  4489--4497. IEEE, 2015.

\bibitem{wang:hal-00725627}
H.~Wang, A.~Kl{\"a}ser, C.~Schmid, and C.-L. Liu.
\newblock {Dense trajectories and motion boundary descriptors for action
  recognition}.
\newblock Research Report RR-8050, Aug. 2012.

\bibitem{wangclassifying}
K.-C. Wang and R.~Zemel.
\newblock Classifying nba offensive plays using neural networks.
\newblock In {\em Sloan Sports Analytics Conference}, 2017.

\bibitem{weinland2011survey}
D.~Weinland, R.~Ronfard, and E.~Boyer.
\newblock A survey of vision-based methods for action representation,
  segmentation and recognition.
\newblock {\em Computer Vision and Image Understanding (CVIU)},
  115(2):224--241, 2011.

\bibitem{zheng2015conditional}
S.~Zheng, S.~Jayasumana, B.~Romera-Paredes, V.~Vineet, Z.~Su, D.~Du, C.~Huang,
  and P.~H. Torr.
\newblock Conditional random fields as recurrent neural networks.
\newblock In {\em Computer Vision and Pattern Recogntion (CVPR)}, pages
  1529--1537, 2015.

\end{thebibliography}
}

\end{document}